\documentclass[letterpaper, 10 pt, conference]{ieeeconf}  
\IEEEoverridecommandlockouts                              
\usepackage[pdftex]{graphicx}
\usepackage{listings}
\usepackage{amsmath,amssymb,amsfonts} 
\usepackage{algorithm,algorithmic}
\usepackage{color}
\usepackage{lipsum}
\usepackage{caption}
\usepackage{comment}
\usepackage{subcaption}
\usepackage{url}           
\usepackage{booktabs}       
\usepackage{algorithm,algorithmic} 

\DeclareMathOperator*{\argmax}{argmax}

\usepackage{tabularx}
\usepackage{cite} 
\usepackage{hyperref}
\usepackage[absolute,overlay]{textpos}

\title{\LARGE \bf
Learning Modular Robot Visual-motor Locomotion Policies
}

\author{Julian Whitman and Howie Choset
\thanks{Carnegie Mellon University, Pittsburgh PA, USA.}%
}

\begin{document}

\maketitle
\thispagestyle{empty}
\pagestyle{empty}

\begin{textblock*}{5in}(0.55 in,0.5in) 
Preprint: To appear in the 2023 International Conference on Robotics and Automation
\end{textblock*}

\begin{abstract}

Control policy learning for modular robot locomotion has previously been limited to proprioceptive feedback and flat terrain. This paper develops policies for modular systems with vision traversing more challenging environments. These modular robots can be reconfigured to form many different designs, where each design needs a controller to function. Though one could create a policy for individual designs and environments, such an approach is not scalable given the wide range of potential designs and environments. To address this challenge, we create a visual-motor policy that can generalize to both new designs and environments. The policy itself is modular, in that it is divided into components, each of which corresponds to a type of module (e.g., a leg, wheel, or body). The policy components can be recombined during training to learn to control multiple designs. We develop a deep reinforcement learning algorithm where visual observations are input to a modular policy interacting with multiple environments at once. We apply this algorithm to train robots with combinations of legs and wheels, then demonstrate the policy controlling real robots climbing stairs and curbs.

\end{abstract}

\section{Introduction}

Vision can help a robot learn to locomote through challenging terrain \cite{agarwal2022legged, yu2021visual, miki2022learning, rudin2022learning }.
Recent work has trained visual-motor control policies, which react in real-time to both exteroceptive (externally-measured/visual) and proprioceptive (internal actuator and IMU) feedback, for legged robots with reinforcement learning (RL) \cite{agarwal2022legged, yu2021visual, miki2022learning, rudin2022learning}.
However, these prior methods must be re-trained any time a different robot mechanism design is introduced, making it computationally expensive to add new designs.
To address this limitation, this paper develops an RL method to train one policy for multiple designs and environments, such that the policy can generalize (i.e., zero-shot transfer, apply without additional optimization/learning) to new designs and environments.

\begin{figure}[tb] 
     \centering
      \includegraphics[width=0.77\linewidth]{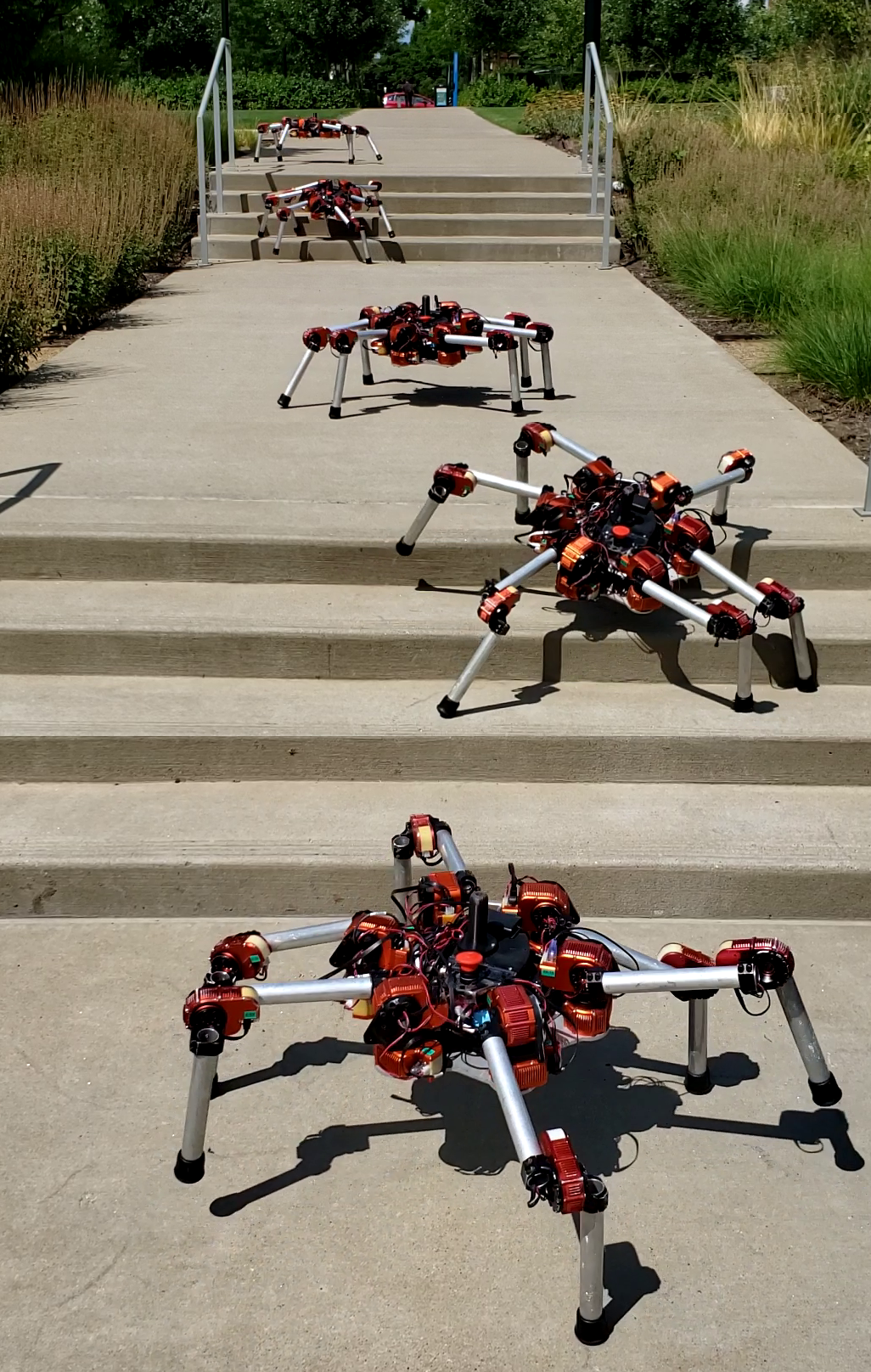}
      \caption{A time-lapse of our hexapod robot climbing stairs using the onboard vision system and our learned modular visual-motor policy. The steps in the stairs shown here are 14 cm high, which at the robot's nominal resting stance, is near the distance between the robot body and the ground. A video of this behavior is in the supplementary material and at \url{https://youtu.be/2tCj34zY6kI}
      }
      \label{fig:stairs}
\end{figure}

We focus on modular robots, where a set of modules, e.g., legs, wheels, and a body, can be easily re-combined to construct many designs. 
But a design on its own is not a functional robot; each new design needs a controller. 
It would be inefficient to create a new control policy from scratch for every new design and environment.
Instead, a policy can be trained to control a variety of designs at once. 
Prior work \cite{whitman2021learning, huang2020smp} found that with a modular learning architecture, one policy can control many designs, and even generalize to new designs outside of those seen in training.
But, those policies operated only in flat obstacle-free environments using proprioceptive sensor feedback.

This paper develops ``Visual-motor MBRL for modular robots,'' an algorithm that builds on our prior learning methods \cite{whitman2021learning} to traverse more challenging terrain. 
This work adds exteroceptive inputs, multiple terrains during training, and alters the RL pipeline to support discovery of climbing behaviors. 
We train one policy with a set of three designs and two environments, and show the policy can generalize to both new designs and environments. 
Then, we investigate how including more than one design and environment during training impacts policy performance after convergence. 
Finally, we instrument our robot with onboard vision, and demonstrate that our policy can control real robots to climb stairs and curbs.

\section{Background}

\subsection{Reinforcement learning}

RL methods collect data from an agent's interaction with its environment to train a policy \cite{ibarz2021train}. 
RL has been used in recent years with deep neural networks to control a variety of locomoting robots, both in simulation and in reality \cite{ibarz2021train, hwangbo2019learning, xie2020learning,  heess2017emergence, hafner2020towards, zhang2021importance, amos2021model}.
Recent work has shown the benefit of including vision as an input to a policy trained with RL, i.e., those that take exteroceptive inputs in addition to proprioceptive inputs over those that take only proprioceptive inputs \cite{agarwal2022legged, yu2021visual, miki2022learning, rudin2022learning}.

RL methods can broadly be categorized as model-free RL (MFRL) or model-based RL (MBRL). 
MFRL treats the robot-environment interactions as a black box, and has been used to train visual-motor policies for quadrupeds \cite{miki2022learning, rudin2022learning} that run in real-time.
In contrast, MBRL explicitly learns an approximate model of the robot-environment dynamics.
MBRL has been shown to be significantly more sample-efficient \cite{nagabandi2018neural, chua2018deep,amos2018differentiable, rajeswaran2020game}, that is, requires fewer trajectories to reach convergence. 
Though most recent MBRL work operates in uniform environments without exteroceptive inputs, \cite{ wang2022rough} added exteroceptive measurements to the model input, and trained on rough terrain.
But, \cite{ wang2022rough} was limited to wheeled robots, and did not produce a real-time reactive policy, instead using the approximate model to optimize control actions off-line.

\subsection{Modular policies}

The \textit{modular policy} learning framework, introduced in \cite{whitman2021learning}, is geared toward systems where a large number of designs are derived from a set of modules. 
Modular policies can learn to control robots composed of various combinations of modules, then transfer without additional training to new robots with different designs.
In these policies, each type of module has a neural network associated with it-- i.e. there is one network used to control all of the ``leg'' modules, and a different one used to control all of the ``wheel'' modules.
The structure of a modular robot is represented as a design graph, with nodes as modules and edges as connections between them. 
A design graph is used to create a policy graph with the same structure.
Each node in the policy graph uses the neural network associated with its corresponding module type's node to convert the inputs from that module's observations into the outputs for that module's actions. 
For example, the part of the neural network for leg-type modules is used to compute actions for each of the legs on a hexapod, one leg at a time.

Modular policies are implemented with a graph neural network (GNN) architecture \cite{scarselli2008graph,wang2018nervenet}, which enables modules to learn to modulate their outputs via a communication procedure in which they send information over the graph edges. 
As a result, the policy can produce different behaviors for the same module depending on its location and neighboring modules within the robot.
In this work, we use the policy architecture of \cite{whitman2021learning}, with additional observations for local terrain measurements included in the body node input.

\section{Modular visual-motor policy learning} \label{sec:mbrl_ext}

\subsection{Problem formulation}

The designs to be trained $D_\textrm{train}$ are first chosen by the user.
The objective we optimize maximizes the reward a policy receives when applied to multiple designs and environments, 
\begin{align} \label{eqn:policy_opt}
     \theta^* =  &\argmax_{\theta} 
    \mathbb{E}_{e \sim \mathcal{E}}  
     \bigg[ \overbrace{ \frac{1}{|D_{\textrm{train}}|} \sum_{d \in D_{\textrm{train}}} \underbrace{\mathbb{E}_{a \sim \pi_\theta} \sum_{t=1}^T r (s_t, a_t, d )}_{
     \substack{\text{Expected reward} \\ \text{for each design}}
      }}^{\text{Average over designs}} \bigg] \\
    & \textrm{s.t.} \quad s_{t+1} = f(s_t, a_t, e, d) \qquad \textrm{(Dynamics)}\\
   & a \sim \pi_\theta(a|o, d) \qquad \textrm{(Policy)} \\
   & o \sim O(s) \qquad \textrm{(Observation function)} \\
   & s_o \sim p(s) \qquad \textrm{(Initial state distribution)} 
\end{align}
Here $\pi_\theta$ is a stochastic policy conditioned on observations and parameters $\theta$. 
The policy is applied to the training designs $D_{\textrm{train}}$ on environments $e$ sampled from a distribution of non-flat environments, $e\sim \mathcal{E}$. 
The reward $r:S\times A \times \mathcal{D} \to \mathbb{R}$ is computed over a time horizon $T$.
The observation function $O$ adds noise to, or removes, parts of the state vector. 
The observation function we use removes the x/y position, yaw, and the linear velocity of the body from the state, as these can be estimated by an odometry system, whereas the remainder of the state (joint positions and velocities, body orientation and angular velocity) can be readily obtained from joint and IMU sensors. 
During training, a low level of Gaussian noise is added to all these observations, with variance tuned to approximate the noise levels of hardware sensors.
Exteroceptive observations (e.g. from a vision system) are incorporated into the observations for the body module, making $\pi$ a visual-motor policy, rather than a purely proprioceptive policy.

This problem formulation is similar to that used in \cite{whitman2021learning}, with the addition of marginalization over multiple environments.  
We approach this problem with an MBRL algorithm, which learns an approximate model $\tilde f_\phi$ with parameters $\phi$, then optimizes a policy with the model as a differentiable approximation of the dynamics.

\subsection{Algorithm overview}

Our algorithm alternates between phases of model learning, policy optimization, and data collection, within each iteration.
This work differs from that of our prior and related MBRL methods because each phase gathers data from, and learns to apply to, multiple designs and environments at once.
The psuedocode of the method is in Algorithm \ref{algo:mbrl}.

In the first phase, an approximate dynamics model is trained with supervised learning from a dataset of randomly generated trajectories in the simulation environment.
Next, the model is used within policy optimization, leveraging the fact that the neural network-based model is differentiable and can be applied in large parallel batches. 
The policy is then applied to robots in simulation to generate more trajectory data. 
The dynamics model is retrained with the additional data, the policy re-optimized, and the process repeated. 
All data is gathered from simulation (NVIDIA IsaacGym \cite{makoviychuk2021isaac}) with a simulation time step of $1/60$ s, and each action is applied for 5 time steps.

\subsection{Initial model data acquisition}
\label{sec:initial_data}
First, the designs $D_\text{train}$ are randomly initialized at near their nominal joint positions on a uniform flat environment. 
Actions are created by sampling $10$ normally distributed random joint commands, and fitting splines to create smooth joint commands over $100$ time steps.
These actions are applied to obtain initial trajectories stored in a dataset $\mathcal{T}_\textrm{train}$.

\subsection{Model learning}

Next, the dataset $\mathcal{T}_\textrm{train}$ is used for supervised model learning (as in \cite{whitman2021learning, yang2020data}). 
The modular model $\tilde f_\phi$ has the same structure as the modular policy, that is, it is a GNN with body, leg, and wheel module node types.
The body node takes as input the state of the body (world frame orientation and velocity). 
In contrast to \cite{whitman2021learning}, we incorporate an additional local terrain observation processed through convolutional layers, flattened, batch-normalized, and appended to the body node input.
The leg and wheel nodes take in joint angles and joint velocities from their respective modules.
Each node outputs a change in state similarly to \cite{whitman2021learning, nagabandi2018neural, sanchez2018graph, yang2020data}.

\subsection{Policy optimization}

The policy is optimized to maximize reward with respect to the model.
Our policy optimization method differs from that of \cite{whitman2021learning}, which used the model to create a dataset of trajectories, then trained a policy via behavioral cloning independently from the model.
This work instead optimizes the policy network parameters end-to-end with the model.
The loss function derived from (\ref{eqn:policy_opt}) is
\begin{equation} \label{eqn:RL_loss}
     \mathcal{L}_{\textrm{RL}} =  - \mathbb{E}_{e \sim \mathcal{E}}
     \bigg[  \frac{1}{|D_{\textrm{train}}|} \sum_{d \in D_{\textrm{train}}} \mathbb{E}_{a \sim \pi_\theta} \sum_{t=1}^T r (\tilde s_t, a_t,  d )  \bigg].
\end{equation}

When optimizing this loss, the approximate state evolves according to the model $\tilde s_{t+1} = \tilde f_\phi(\tilde s_t, a_t, e, d)$, where the model is held fixed during policy optimization.
The actions are sampled according to $a_t \sim  \pi_\theta(a_t|o_t, d)$, $o_t \sim O(\tilde s_t)$.
The initial state $\tilde s_t = s_0$ is taken from the simulator.
Over the time horizon length $T$, the policy is applied to the approximate state, a reward is computed, and the state is advanced according to the model.
The total reward over $T$ steps is accumulated in $\mathcal{L}_{\textrm{RL}}$, and gradients $d\mathcal{L}_{\textrm{RL}}/d\theta$ are used in gradient descent with an Adam optimizer \cite{kingma2014adam}. 
In other words, after a fully differentiable $T$-step roll-out of forward passes using the model and policy, policy parameter gradients are computed by back-propagating through the sequence of states and actions.
This process is repeated with random mini-batches of initial states. 
We also use a one-step ($1/12$ s) simulated latency \cite{simtoreal2018} to facilitate transfer to reality.

The reward function $r$ is designed to reward a variety of robot designs for locomoting forward over rough terrain.
The main term in the objective rewards the distance in the $+x$ direction. Additional terms with smaller relative weights penalize roll, pitch, yaw, deviation from $y=0$, control effort, and distance from a nominal stance. 
The exteroceptive observations are taken by sampling the simulation environment height at the approximated state's position. 
The policy learns locomotive behaviors that vary depending on the design and on the environment sensed.

\subsection{On-policy model data acquisition}

After the policy training phase, the policy is used to gather additional trajectories in simulation.
In early iterations, the model may be inaccurate, such that the policy optimization can exploit model bias to produce a policy that will be low-cost under the model but could be high-cost in the simulation environment \cite{deisenroth2011pilco}. Consequently, policy optimization can become unstable if the time horizon is too long initially. 
To iteratively reduce model bias, we apply the policy to gather trajectories, and fine-tune the model with this new data.
At the next iteration of policy optimization, the model will be more accurate near the policy, causing a virtuous cycle in which the model improves near states visited by the policy, and the policy is optimized with respect to the refined model.

The policy is applied to the simulation environment deterministically using the mean output of the policy distribution $\pi_{\theta}$.
The resulting states and actions are added to $\mathcal{T}_\textrm{train}$.
The policy can also be applied to the simulation environment with time-correlated noise.
That is, the policy is applied to the model for $T$ steps, $x_{t+1} = \tilde f_\phi(x_t, \pi_{\theta}(o_t, e_t))$, to obtain control actions $u_{1:T}$. 
These control actions are perturbed with time correlated noise similarly to the actions generated in Sec. \ref{sec:initial_data}, then applied to the simulation.
Trajectories across the designs and environments  are added to $\mathcal{T}_\textrm{train}$. 
The model is then refit using supervised learning, and used for the next phase policy optimization.

\subsection{Terrain curriculum}

We created an adaptive curriculum on the terrain difficulty, inspired by
\cite{rudin2022learning, miki2022learning}.
At the first iteration, the terrain is fully flat, presenting the easiest learning problem. 
The model and policy are trained initially on flat ground, and once all designs pass a threshold distance travelled, the height of terrain features in all environments is incrementally increased by 2 cm. 
Then, each batch in the policy training and on-policy trajectories contains some samples from each terrain difficulty level to prevent catastrophic forgetting.

\subsection{Additional implementation details}

We found that for our algorithm to succeed, a few additional training implementation details were helpful.

The recurrent policy trained more reliably than a non-recurrent policy; a potential cause for this could be vanishing gradients over the time sequence in which the policy loss was computed. 
We train the recurrent policy to operate for more than $T$ time steps (the policy optimization horizon) with truncated back-propogation through time \cite{werbos1990backpropagation}. 
Each time a batch of states are sampled as initial states for the policy rollout, half of those states come from the simulation of the previous policy iteration, and half of those states come from an ``imagination'' of applying the policy to the current model. 
The states that come from imagination are associated with a policy hidden recurrent vector, used as the initial hidden recurrent vector for those initial states.
Using a policy that is recurrent over time, and not just recurrent over the internal propagation phase of each GNN forward pass, has an additional benefit.
We were able to reduce the number of internal propagation steps within the GNN at each time step to only one step.
As a consequence, the forward pass is slightly less computationally expensive than it would be for greater number of internal propagation steps  (see \cite{wang2018nervenet, whitman2021learning} for descriptions of GNN internal propagation steps). 

Secondly, we found that with a fixed learning rate, the policy optimization would sometimes diverge. 
We implemented a simple adaptive learning rate to stabilize policy optimization.
At the start of policy optimization, we compute the net reward from applying the policy for $T$ steps to a large set (e.g. 10x the batch size) of initial states. 
This ``validation reward'' acts similarly to a validation loss in a supervised learning problem. 
If the current validation reward becomes lower than that initial validation reward, we revert the policy parameters back to the last point when the validation reward was computed, lower the learning rate, and continue.

Thirdly, we found regulating the stochastic policy entropy, by rewarding the variance  output by the policy, aids in balancing exploration and exploitation.
During policy optimization, a small entropy bonus is added to the RL loss (as in \cite{haarnoja2018soft}), which prevents premature convergence to a poor local minima and aids in discovering the transition from flat-ground to climbing behavior.

\begin{figure}[tb] 
     \centering
     \vspace{0.4em}
     \includegraphics[width=\linewidth]{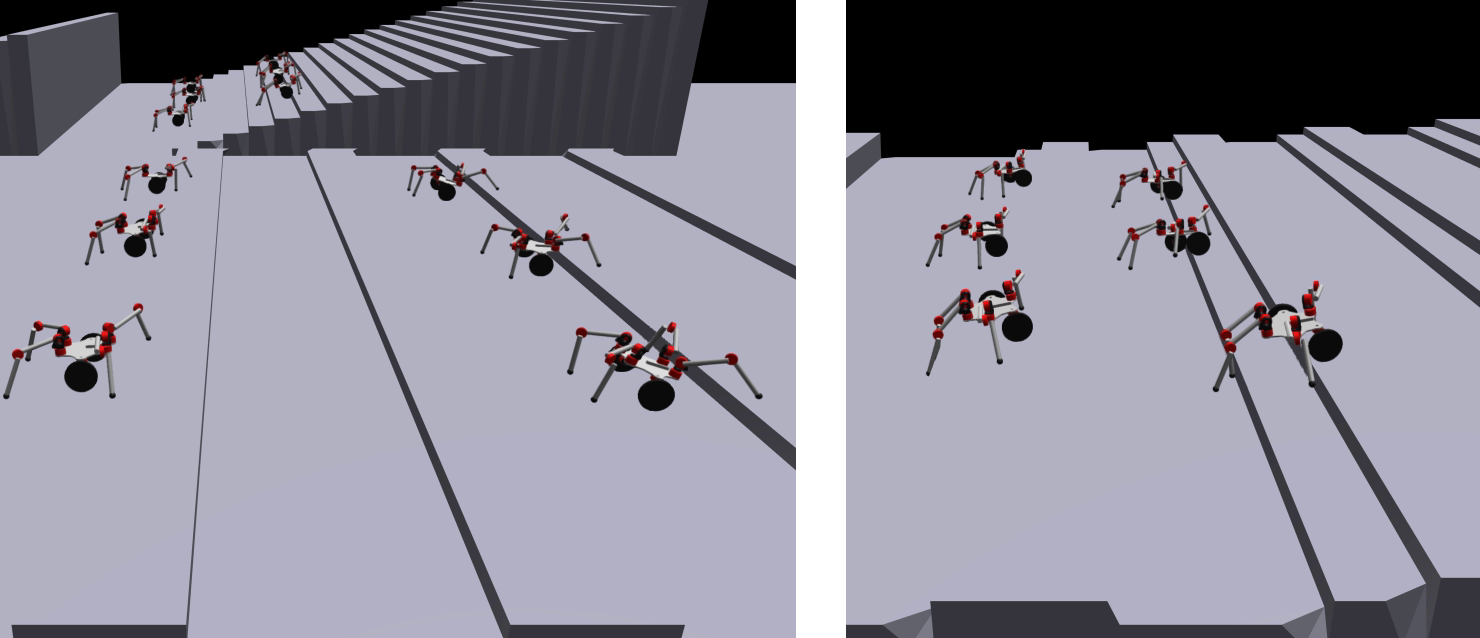}
      \caption{Our modular visual-motor policy is trained simultaneously using multiple designs and environments at the same time. Left: One of the designs trained to locomote through stairs and curbs. Right: Generalization to a new design and a new environment not seen during training.
      We show instances of the same robot at two time steps, at states reached from three initial conditions.
      }
      \label{fig:sim}
\end{figure}

\begin{algorithm}[t]

 \caption{Visual-motor MBRL for modular robots. Each step is conducted for multiple designs and environments.  
 }
 \label{algo:mbrl}
 \begin{algorithmic}[1]

  \STATE Collect dataset $\mathcal{T}_\textrm{train}$ from random action trajectories
  \FOR {$i = 1 \dots N$}
    \vspace{0.5em}\STATE \textit{Model learning phase:}
  \STATE Train model $\tilde f_\phi$ from $\mathcal{T}_\textrm{train}$ with supervised learning
    \vspace{0.5em}\STATE \textit{Policy optimization phase:}
    \STATE Initialize buffer $\mathcal{B}$ with random initial robot states and zero-valued policy hidden-state vectors, i.e., $\mathcal{B} \leftarrow \{ s_{0,i}, h_0 \}_{i=1}^{N_\textrm{batch}}$ where $h_0 = \Vec{0}$
      \FOR {$k = 1 \dots K$}
        \STATE Sample a batch of initial states $s_0$ with hidden-state vectors from $\mathcal{B}$, and set the hidden state of the policy $\pi_\theta$
        \STATE $R \leftarrow 0$, $\tilde s_0 = s_0$
        \vspace{0.5em}\STATE \textit{Roll out policy with approx. dynamics}
        \FOR{$t = 1 \dots T$ }
            \STATE $a_t \sim \pi_\theta(a_t|O(\tilde s_t))$, $\tilde s_{t+1} = \tilde f_\phi(\tilde s_t, a_t)$
            \STATE $R \leftarrow R+ r(\tilde s_t, a_t)$
            \ENDFOR
        \STATE $\mathcal{L}_{\textrm{RL}} = -R$
        \STATE Gradient descent on policy parameters $d\mathcal{L}_{\textrm{RL}}/d\theta$

        \STATE Overwrite part of $\mathcal{B}$ with intermediate states and policy hidden-state vectors, i.e., $\mathcal{B}^{N_\textrm{batch}/2 : N_\textrm{batch}} \leftarrow \{ s_{T/2,i}, h_{T/2} \}_{i=N_\textrm{batch}/2}^{N_\textrm{batch}}$
      \ENDFOR

    \vspace{0.5em}\STATE \textit{Terrain curriculum:} Increase environment difficulty for designs that have met the performance threshold
    \vspace{0.5em}\STATE \textit{On-policy model data acquisition:}
    \STATE Apply $\pi_\theta$ to collect data $\mathcal{T}_{\textrm{new}}$
    \STATE $\mathcal{T}_{\textrm{train}}  \leftarrow \mathcal{T}_{\textrm{train}} \cup \mathcal{T}_{\textrm{new}} $

  \ENDFOR
 \end{algorithmic} 
 \end{algorithm}
 
\section{Generalization experiments}

We next measure the policy's capability to generalize to new designs and new environments.
We created a training and test set of designs and environments, a baseline behavior as a basis of comparison, and compared the performance under conditions both seen and unseen in training.

\subsection{Designs and environments tested}

In these experiments, we limit the training designs $D_\textrm{train}$ to three designs, each with four legs and two wheels, where the position of the wheels is left-right symmetric and either in the front, middle, or rear of the body.
The test designs $D_\textrm{test}$ consists of three designs with four legs and two wheels, where the location of the wheels is left-right asymmetric.

We created training environments with two types of terrain.
The first, ``stairs,'' contains ascending steps.
The second, ``curbs'' contains rectangular blocks with fixed width at a regular interval.
The test environment consists of steps and flat regions staggered. Fig. \ref{fig:sim} depicts a visualization of these environments.
The terrain difficulty curriculum includes  ``levels'' of difficulty, ranging from fully flat to higher obstacles, with the maximum step height increasing in 2 cm intervals.
For the following experiments, we report the results of each policy tested on a nearly-flat terrain (2 cm steps) and a more difficult terrain (10 cm steps).

After training, for each design/environment combination, the policy was simulated from 10 perturbed initial conditions. The mean and standard deviation of the distance travelled after 200 time steps (16.7 seconds) was recorded.
Each policy was trained three separate times, and the results averaged for each cell.
Although the reward includes multiple terms, we use the distance travelled forward as a metric for policy success, as the reward is dominated by this term and it is more easily interpretable than the full reward value. 
Training was run for 20 iterations, such that the three-design two-environment process trained for about 8.5 hours on a computer with an 8-core AMD Ryzen 7 2700X processor and a NVIDIA GTX 1070 GPU.

\subsection{Comparison to hand-crafted baseline}

We developed a hand-crafted gait applicable to the various combinations of legs and wheels tested in this work.
All legs are given an alternating tripod gait, with phase and position offsets assigned as if they were in a hexapod.
All wheels are given differential drive/skid-steering commands, with position offsets and wheel speeds assigned as if they were on a car.
Gait speeds and amplitudes were tuned to produce steps as fast and high as could be tracked by the joint velocity limits.
This baseline gait is not fully open-loop, as it reacts to steer to face forwards using the observed yaw angle.  
The baseline gait enables the tested designs to locomote effectively on flat ground and over small obstacles, but its performance degrades as the terrain features (steps, curbs) become larger.
In the tables following, we denote the results from this gait as ``tripod baseline.''

\subsection{Generalization to new designs and environments}

First we test the capability of the policy when applied to new (i.e., ``test,'' unseen in training) designs, new environments, and both simultaneously.
Here the policy is trained with three designs and two environments jointly, then the average distance travelled is recorded in Table \ref{table:1}.
We found that the policy produced larger displacement than the baseline.
These results indicate that the policy can generalize to new designs in new environments at the same time, and still produce behaviors better than our hand-crafted policy, though there is a small drop in performance between the train and test results for the 10 cm steps.

\begin{table}[h]
\centering
\begin{tabular}{|c | c | c |} 
  \multicolumn{1}{c }{}  & \multicolumn{2}{c }{Avg. dist. traveled in 200 time steps (m) } \\
\cline{2-3}
\multicolumn{1}{c |}{2 cm steps}  & Tripod baseline & \multicolumn{1}{c |}{Policy}   \\
  \hline
 Train designs + train envs. & 6.3 & \textbf{7.0} $\pm$ 0.5\\
Test designs + train envs. & 5.9 &\textbf{7.0} $\pm$ 0.4 \\
Train designs + test env. & 6.6 &\textbf{7.1}  $\pm$ 0.3 \\
Test designs + test env. & 6.2 &\textbf{7.3}  $\pm$ 0.3 \\
  \hline
 \multicolumn{1}{c |}{10 cm steps}  & Tripod baseline & \multicolumn{1}{c |}{Policy}   \\
  \hline
Train designs + train envs. &2.3 &\textbf{4.9}  $\pm$ 0.2 \\
Test designs + train envs. &3.5 &\textbf{4.2}  $\pm$ 0.4\\
Train designs + test env. &2.5 &\textbf{4.2}  $\pm$ 1.0 \\
Test designs + test env. &3.6 &\textbf{4.2}  $\pm$ 0.6 \\
   \hline
\end{tabular}
\caption{Generalization to new designs and/or environments. ``Train'' indicates that the design/environment was seen during training, ``test'' indicates that it was not.
The policy mean and standard deviation are listed after the policy is trained from three times with random initial seeds.
The policy was able to pass the baseline in all cases, even when both the designs and environments were not seen during training.}
\label{table:1}
\end{table}

\subsection{The trade-off between specialization and generalization}
Training with multiple designs/environments enables generalization to new designs/environments, but requires optimizing a more complex loss.
We next test whether including multiple designs and/or environments in training effects the policy's performance. 

The policy is trained with one/multiple designs and one/multiple environments at a time. 
When one design/environment at a time is used, then we train policies separately for each, and average the performance of those policies on the conditions in which they were trained.
For example, in the ``Design ind. + Envs. joint'' condition, we train three separate policies, one for each of the three designs. 
Each policy is trained with multiple environments, and other hyperparameters are held constant. 
We then measure the performance of those three policies on their corresponding designs, where each policy is  applied to the environments, and the performance over all designs and environments is averaged. 
When multiple designs/environments are included in training, we also test whether that policy can generalize to test designs/environments. 
Results, averaged over three training runs from random initial seeds, are recorded in Table \ref{table:2}.

\begin{table*}[ht]
\centering
\vspace{0.4em}
\begin{tabularx}{\textwidth}{|l | X | X | X | X | X|} 
\multicolumn{1}{c }{} &  \multicolumn{5}{c }{Avg. dist. traveled in 200 time steps (m)}  \\
\cline{2-6}
  \multicolumn{1}{c |}{} & Tripod baseline & Designs ind. + Envs. ind. & Designs joint + Envs. ind. & Designs ind. + Envs. joint & Design joint + Envs. joint   \\
  \hline
  \hline
  Train designs + train envs. (2 cm) &6.3 &7.2 &7.0 & 7.6 &7.0 \\
  \hline
  Test designs + train envs.  (2 cm) &5.9 &-- &7.0 & -- &7.0  \\
  \hline
  Train designs + test env.  (2 cm) & 6.6 &-- & -- &7.1 & 7.1\\
   \hline
  \hline
  Train designs + train envs.  (10 cm) &2.3 &6.4 &5.1 & 6.2 &4.9  \\
  \hline
  Test designs + train envs.  (10 cm) &3.5 &-- &4.6  & -- &4.2  \\
   \hline
  Train designs + test env.  (10 cm) &2.5 &-- &-- &4.2 & 4.2\\
   \hline
\end{tabularx}
\caption{The effect of including multiple designs and/or environments during training (``Train'' conditions) on policy performance and generalization (``Test'' conditions).
When designs/environments are trained individually, we do not test generalization to unseen designs/environments, and mark these conditions with a ``--''.
} 
\label{table:2}
\end{table*}

A policy trained with multiple designs and environments comes with drawbacks alongside its benefits.
On flat ground and low obstacles, the policy performance was not significantly affected by additional designs or environments in training.
But as the task becomes more difficult with larger obstacles (higher steps), the policy performance was lower when trained with multiple designs and environments than when trained individually for each of those same designs or environments.
We call this the ``joint learning delta:'' when training with more than one design or environment, the performance on the more difficult terrain decreased.

The joint learning delta is most apparent from the difference between the ``Designs ind. + Envs. ind.'' condition and the ``Designs joint + Envs. joint'' condition in Table \ref{table:2} 
The policy is fit to a more complex objective, rather than specializing to a single design and environment. 
As a consequence, though it gains the ability to generalize, we observed a decreased performance compared to a specialized policy. 
Jointly training with multiple designs and environments will likely require larger neural network capacities, larger batch sizes, and more training time in order to reach the performance of the policy that is trained with a single design and environment.

\section{Robot demonstrations}

Given that ``simulations are doomed to succeed'' \cite{brooks1993real}, we validated our policies in  reality.
We constructed a modular robot with onboard power (GRIN batteries \cite{GrinWebsite}), computation (4-core i7 Intel NUC), sensing (a Realsense D435 and XSens IMU), and Hebi X-series actuators \cite{HebiRoboticsWebsite}.
 The onboard vision system uses VINS-Fusion \cite{qin2017vins} for odometry, and Elevation Mapping \cite{Fankhauser2018ProbabilisticTerrainMapping, Fankhauser2014RobotCentricElevationMapping} to produce maps. 
 A local terrain map of $21 \times 21$ grid of points (1.5 by 1.5 m) aligned with the robot body frame is sent as input to the policy at each time step. 
Note this robot does not have foot contact sensing, so it cannot directly sense ground contacts.
 
The policy outputs target joint velocity commands tracked by low-level PID loops at a higher frequency on-board the actuator.
The actuators perform more accurate tracking when provided with a feed-forward (FF) torque value, in particular when they are under heavy load.
Therefore in addition to control outputs, we train a torque estimation network (implemented as a GNN with the same structure as the policy) that estimates the feed-forward torque needed for the actuator to track the desired velocity.
The data for this training is obtained by tracking the torques experienced in simulation.

Our policy was able to transfer from simulation to reality.
Most notably, the policy was able to control a hexapod to climb outdoors human-scale stairs. 
These demonstrations are shown in Figures \ref{fig:stairs} and \ref{fig:curb}, in the supplementary video and at \url{https://youtu.be/2tCj34zY6kI}.

To test the reliability of the simulation-to-reality transfer, we conducted trials of the policy applied to a hexapod traversing a 19 cm-high roadside curb.
The robot was reset one meter from the curb, then the modular visual-motor policy applied for 20 s. 
We conducted five trials, and in each one, the robot was able to locomote over the curb, transitioning from a flat-ground walking motion to a climbing motion and back to flat ground. 
We also applied the baseline alternating tripod behavior, with the step size high enough to step on to the curb. 
Though effective on flat ground, the alternating tripod baseline only traversed the curb one out of five trials, indicating that this obstacle is difficult enough to require a different behavior to succeed.
Finally, we conducted a test in which the visual-motor policy at run-time was given a spoofed terrain map image consistent with walking on flat ground, effectively ``blindfolding'' the policy.
The ablated policy locomoted on flat ground, but did not traverse the curb in any of five trials, indicating it uses the exteroceptive feedback to adapt its behavior.

\begin{figure}[tb] 
\vspace{-0.5em}
     \centering
      \includegraphics[height=0.22\linewidth]{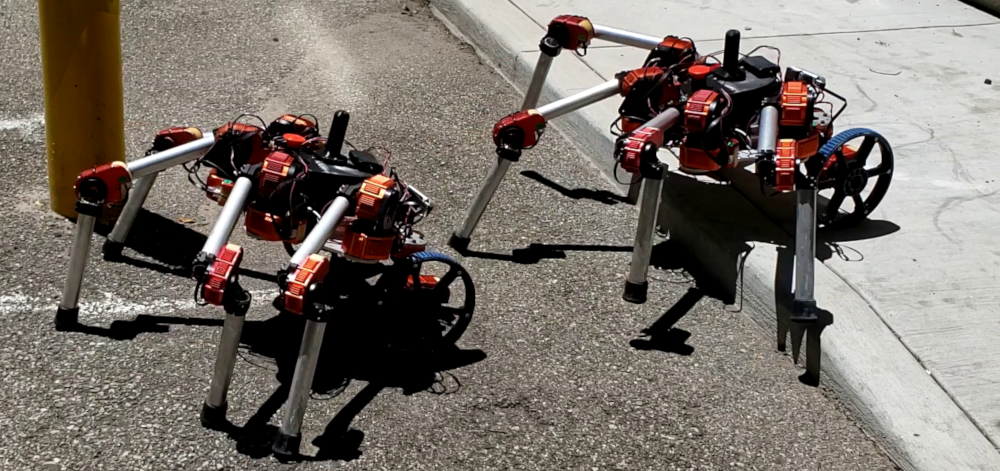}
      \includegraphics[height=0.22\linewidth]{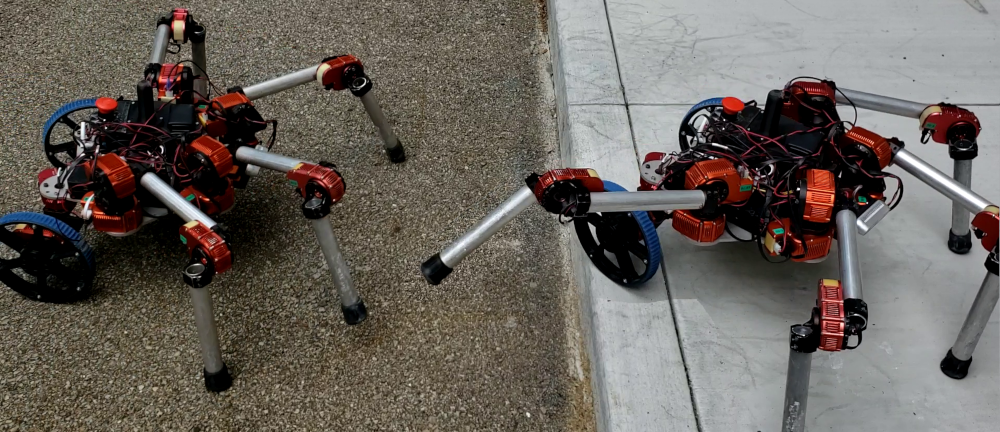}
      \includegraphics[height=0.22\linewidth]{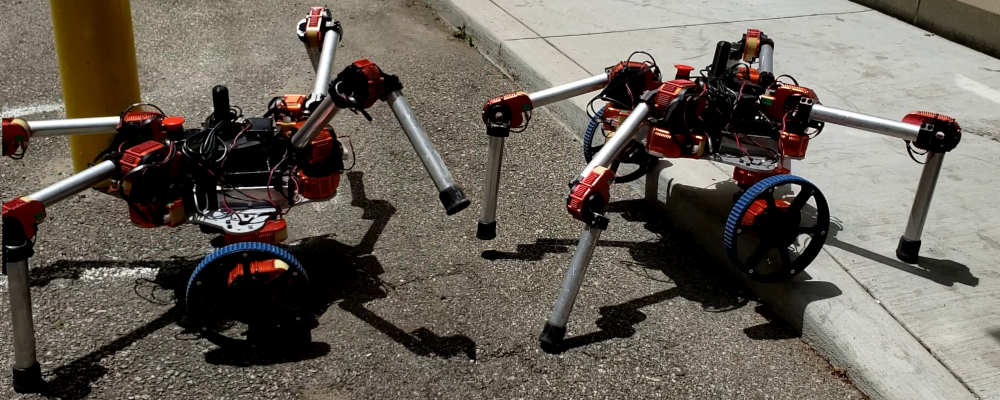}
      \caption{Three time-lapses of robots with four legs and two wheels climbing up a curb using the onboard vision system and modular visual-motor policy.
      }
      \label{fig:curb}
\end{figure} 

\section{Conclusions} \label{sec:conclusion}

This work presents a deep RL algorithm that trains a modular visual-motor policy to control multiple designs in multiple environments.
We showed that the policy can generalize to new designs and environments, and transfer from simulation to reality.
To the best of our knowledge, this is the first demonstration of a legged robot locomoting through non-flat terrain using  MBRL.

These methods also come with some limitations. 
We validated that our policy can transfer from simulation to reality, but on more difficult environments, the transferred behaviors degrade compared to their simulated counterparts, though we have yet to quantify the sim-to-real gap for the full variety of designs and environments used in this work. 
Future work will aim to adapt techniques such as domain randomization \cite{wang2021domain} or domain adaptation \cite{kumar2021rma}. 
More accurate robot models in simulation, or learning from a mix of simulation and real data, may also improve sim-to-real.
Further, we showed only a limited number of designs and environments, and ongoing work aims to increase the variety of designs and environments included.

Another direction for future work is to learn partially from expert knowledge. 
This paper could be combined with our concurrent work on combining imitation and reinforcement learning for modular robots \cite{whitman2022learning}, such that the climbing policy could either be warm-started by imitating an existing policy or hand-engineered gait, even when the demonstrations are shown for designs not contained within $D_\textrm{train}$.

We developed an MBRL method to train modular policies, but similar concepts could be applied to MFRL methods. 
Due to our focus on generalizability in this work, we have not yet directly compared how well our policy performs relative to MFRL methods,  those applied to individual designs in related work, or with different policy architectures.
A direction for future work is adapting methods such as \cite{rudin2022learning} or \cite{huang2020smp}, to modular robots with vision in varied environments.

\bibliographystyle{IEEEtran}
\bibliography{main}

\end{document}